\pdfoutput=1

\documentclass[11pt]{article}

\usepackage[final]{acl}
\usepackage{times}
\usepackage{latexsym}
\usepackage{comment}
\usepackage{algorithm}
\usepackage{algpseudocode}

\usepackage[T1]{fontenc}

\usepackage[utf8]{inputenc}

\usepackage{microtype}

\usepackage{inconsolata}

\usepackage{graphicx}
\usepackage{latexsym}
\usepackage{amssymb}
\usepackage{amsmath}
\usepackage{amsthm}
\usepackage{booktabs}
\usepackage{enumitem}
\usepackage{graphicx}
\usepackage{color}
\usepackage{multirow}
\title{Spec-VLA: Speculative Decoding for Vision-Language-Action Models with Relaxed Acceptance}

\usepackage{ifsym}
\usepackage{footmisc}
\DefineFNsymbols*{newfootnote}{%
    {\textasteriskcentered}
    {\dag}
    {\S}
    \P{**}
    {\ding{172}}{\ding{173}}{\ding{174}}}
\setfnsymbol{newfootnote}

\newcommand*\samethanks[1][\value{footnote}]
{\footnotemark[#1]}
\author{
Songsheng Wang$^{1}$\thanks{Equal contribution.} \thanks{Work was done during the internship at NICS-EFC Lab, Department of Electronic Engineering, Tsinghua University.} \and
Rucheng Yu$^2$\samethanks[1] \and 
Zhihang Yuan$^2$ \and \\
\bf{Chao Yu}$^3$$^4$\and 
\bf{Feng Gao}$^3$\and 
\bf{Yu Wang}$^3$ \and 
\bf{Derek F. Wong}$^1$\thanks{Corresponding author.} \\
$^1$ NLP$^2$CT Lab, Department of Computer and Information Science, University of Macau \\
$^2$ Infinigence AI
$^3$ Tsinghua University
$^4$ Zhongguancun Academy \\
\{\texttt{nlp2ct.songsheng,ruchengyu1130,hahnyuan}\}@gmail.com\\
\{\texttt{yuchao,yu-wang}\}@mail.tsinghua.edu.cn\\
gaof22@mails.tsinghua.edu.cn\\
derekfw@um.edu.mo\\
}

\begin{document}
\maketitle
\begin{abstract}
Vision-Language-Action~(VLA) models have made substantial progress by leveraging the robust capabilities of Visual Language Models (VLMs). However, VLMs' significant parameter size and autoregressive (AR) decoding nature impose considerable computational demands on VLA models. While Speculative Decoding (SD) has shown efficacy in accelerating Large Language Models (LLMs) by incorporating efficient drafting and parallel verification, allowing multiple tokens to be generated in one forward pass, its application to VLA models remains unexplored. This work introduces Spec-VLA, an SD framework designed to accelerate VLA models. Due to the difficulty of the action prediction task and the greedy decoding mechanism of the VLA models, the direct application of the advanced SD framework to the VLA prediction task yields a minor speed improvement. To boost the generation speed, we propose an effective mechanism to relax acceptance utilizing the relative distances represented by the action tokens of the VLA model. Empirical results across diverse test scenarios affirm the effectiveness of the Spec-VLA framework, and further analysis substantiates the impact of our proposed strategies, which enhance the acceptance length by $44\%$, achieving $1.42\times$ speedup compared with the OpenVLA baseline, without compromising the success rate. The success of the Spec-VLA framework highlights the potential for broader application of speculative execution in VLA prediction scenarios. We make our code and data publicly available at \url{https://github.com/PineTreeWss/SpecVLA}.
\end{abstract}

\section{Introduction}
The Vision-Language-Action (VLA) models~\cite{brohan2022rt,brohan2023rt,mees2024octo,wuunleashing, cheang2024gr,vuong2023open} have achieved significant progress by leveraging the rich understanding and generation capabilities from pre-trained visual encoders or Visual Language Models (VLMs). These models can generate robot actions following language instructions. With the development of large-scale robot prediction datasets, recently proposed VLA models such as OpenVLA~\cite{kim2024openvla} demonstrate high generalizability across diverse tasks and environments~\cite{li2024robonurse}.

To achieve the goals above, the parameter size of backbone VLMs is substantial, increasing the computational demand for robot control systems. Meanwhile, the VLMs' Autoregressive (AR) next-token-prediction strategy further increases the decoding latency of VLA models. A series of studies address the efficiency issue through model architecture redesign~\cite{wen2025tinyvla,liu2024robomamba} or task-specific optimizations~\cite{kim2025fine}. Other efforts incorporate Large Language Model (LLM) inference acceleration methods such as Early-Exit~\cite{schuster2022confident} and Jacobi-Decoding~\cite{kou2024cllms} into VLA inference~\cite{yue2024deer,song2025accelerating}. However, incorporating such methods requires resource-intensive fine-tuning of the backbone VLM for Early-Exit~\cite{yue2024deer} or pretraining for Jacobi-Decoding~\cite{song2025accelerating}. 
Moreover, in Jacobi-Decoding, enabling parallel decoding degrades the model performance compared to AR decoding~\cite{song2025accelerating}. 

\begin{figure}[t]
\centering
\includegraphics[width=1\columnwidth]{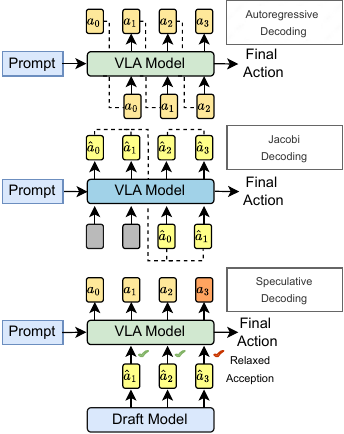}
\caption{Comparison of Autoregressive Decoding, Jacobi-Decoding, and Spec-VLA Decoding Framework. Spec-VLA framework enables parallel generation without tuning or retraining for target VLA model.}
\label{fig:eurai}
\end{figure}

Speculative Decoding ~(SD) provides a lossless solution and also allows for the parallel generation of LLMs. A typical SD architecture, such as Eagle~\cite{li2024eagle_o}, employs a draft model to generate draft tokens efficiently, with the LLMs serving as the verification model to ensure the correctness of these tokens. As the parameters of the draft model are decoupled, additional fine-tuning of the verification model is not required.

Recent works have applied the SD framework in visual generation~\cite{jang2024lantern,park2025lantern++}, designing a task-specific methodology to relax the acceptance for the verification model. The application of SD architecture to accelerate the VLA prediction task is intuitive, enabling efficient adaptation to speed up the generation of downstream tasks while retaining the knowledge of the VLA backbone model. However, its application to the VLA model has not yet been explored.

This work introduces the speculative decoding framework to the AR robot action generation. We propose the Spec-VLA, the first SD framework designed for VLA inference acceleration, which applies the advanced features of the speculative decoding to the robot action generation scenarios.
Surprisingly, the direct application of the SD framework yields minor speed improvements due to the intricate difficulty of VLA prediction for the draft model and the greedy decoding strategy. To further boost the generation speed, we propose utilizing VLA models' token representation to relax the acceptance based on the action distance between draft tokens and ground-truth tokens. Empirical results across various test scenarios demonstrate the effectiveness of the Spec-VLA framework, enabling an acceptance length from 2.10 to 2.94. Analysis confirms that our proposed relaxation of acceptance strategy significantly enhances the acceptance length by $26\%$ to $44\%$, enhances the generation speed by $1.22\times$ to $1.42\times$ while maintaining the success rate of the VLA models. 

Existing reinforcement learning studies have highlighted the importance of robustness, either in multi-task settings where policy degradation occurs ~\cite{Bai_Zhang_Tao_Wu_Wang_Xu_2023,Liu_Gao_Wei_2025}, or under adversarial perturbations where agents exhibit vulnerability ~\cite{Bai_Liu_Du_Wen_Yang_2025}. Inspired by these studies, we further analyze the robustness of VLA models under speculative decoding with relaxed acceptance by exploring the relaxation threshold under multiple tasks, demonstrating the potential of SD frameworks in the VLA prediction domain.
\section{Related Works}
\subsection{Acceleration for VLA Models}
Recent advances in accelerating VLA models can be broadly categorized into multiple directions.
Token-level optimization methods reduce computational redundancy through vision-language token selection. The FastV~\cite{pertsch2025fast} distills task-relevant visual features using auxiliary transformers, while SparseVLM ~\cite{zhang2024sparsevlm} dynamically prunes tokens via spatial attention thresholds. Though efficient without architectural changes, these approaches rely heavily on heuristic token selection, risking generalization failures in novel scenarios.

Conventional LLM acceleration techniques like quantization, pruning, and early-exit strategies have also been adapted for VLA scenarios. QAIL ~\cite{park2024quantization} employs quantization-aware fine-tuning but suffers from precision loss. Mope-CLIP~\cite{lin2024mope} explores modality-specific pruning for vision-language models,  and DeeR~\cite{yue2024deer} implements early-exit mechanisms that compromise action trajectory coherence. While effective in constrained settings, such methods often degrade cross-modal interaction quality and require task-specific tuning.

Structural modifications, such as Robomamba~\cite{liu2024robomamba} and TinyVLA~\cite{wen2025tinyvla}, redesign model backbones using lightweight SSM or distilled vision encoders, achieving latency reduction through structural simplification. The ~\citet{kim2025fine} propose temporal consistency losses to regularize action smoothness, and~\citet{song2025accelerating} reformulate decoding via Jacobi iteration for parallel trajectory generation. The aforementioned methodologies not only require domain-specific data fine-tuning or retraining but also introduce augmented system complexity through model architectural redesign.
Beyond architectural redesign and decoding strategies, recent RL-based studies also target efficiency in VLA inference. 
SEER~\cite{bai2024efficient} improves sample efficiency via aligned experience estimation and policy regularization, 
while D3P~\cite{Yu_Gao_Wu_Yu_Wang_2025} accelerates inference by adaptively adjusting diffusion steps. 
Together, they highlight complementary RL-driven strategies for efficient VLA inference.

\subsection{Speculative Decoding for LLMs}
The SD has emerged as an effective paradigm for inference acceleration in AR generative models, such as machine translation models~\cite{stern2018blockwise} and decoder-only LLMs~\cite{chen2023accelerating}.
 The evolutionary trajectory of SD frameworks reveals three distinct development phases. Pioneering SD frameworks exemplified by Medusa~\cite{cai2024medusa} and Medusa-CTC~\cite{wen2024speculative} introduced parallel generation capabilities through multi-head decoding architectures coupled with tree-attention verification mechanisms. Subsequent developments in the Eagle series, including Eagle~\cite{zhang2024eagle} and Eagle-2~\cite{li2024eagle}, advanced the paradigm through architectural innovations in draft modeling, achieving superior speedup ratios via high-quality draft token generation.  Recently, the Eagle-3~\cite{li2025eagle} and HASS~\cite{zhang2024learning} have further improved the generation capabilities by employing a training-time testing strategy. The framework have shown remakable superiority for LLM acclleration~($3.2\times$ - $5.6\times$), compared with Jacobi-Decoding~($2.5\times$ - $3.0\times$)~\cite{kou2024cllms} and Early-Exit Decoding~($1.9\times$ - $1.8\times$)~\cite{liu2024kangaroo}.

 Recent works have further extended SD applications to emerging scenarios, including retrieval-argumented generation~\cite{wang2024speculativerag} and long-context generation~\cite{yang2025longspec}.   However, empirical validation remains insufficient for multimodal generation contexts. Initial investigations by~\citet{jang2024lantern} demonstrated significant performance degradation when applying existing SD frameworks to visual AR generation tasks. ~\citet{gagrani2024speculative} conducted systematic analyses of visual feature utilization in multimodal applications such as visual question answering and image captioning. Despite these advances, the application of SD methodologies within the VLA generation scenario remains unexplored. 
 
 Relaxed acceptance proves effective in the SD framework, demonstrating particular promise for extending efficiency gains to novel application scenarios. It boosts throughput by loosening the criteria for accepting proposed tokens, striking a balance between efficiency and fidelity. Spec-Dec~\cite{xia-ge-wang-etal-2022} replaces the strict greedy check by accepting any drafted token appearing in the AR model’s top-k candidates, significantly raising token acceptance rates and overall throughput without degrading output quality. Meanwhile, the Lantern framework~\cite{jang2024lantern} accepts the top-k similar tokens in the dictionary, which significantly boosts the generation speed for visual generation. ~(Further improvements~\cite{li-chen-holtzman-etal-2024,zhang2023draft}). These advancements proves the potential of relaxed acceptance in enhancing the efficiency of multimodal models, such as VLA models.
\section{Background}
\subsection{Decoding of VLA Models}
Large VLA models~\cite{ma2024survey}, such as OpenVLA~\cite{kim2024openvla} and RT-2~\cite{brohan2023rt} series, predict action sequences to control robots. They employ a sequence of action tokens $A=\{a_0,...,a_L\}$ which represent the actions at each timestep. Using VLM inference, the model autoregressively predicts seven action tokens to define a control action, including ``$\Delta\text{pos}_x$'', ``$\Delta\text{pos}_y$'' ,``$\Delta\text{pos}_z$'', ``$\Delta\text{rot}_x$'' , ``$\Delta\text{rot}_y$'', ``$\Delta\text{rot}_z$'' and ``gripper\_extension''. Specifically, they utilize greedy decoding, predicting the most probable action token $a_i$ based on the previously predicted tokens $a_{0:i-1}$, visual observations $o$, language instruction prompts $p$, and the learnable model parameters $\theta$.
\begin{align}
    a_i &= \mathop{\operatorname{argmax}}\limits_{a_i} \left[ P(a_i \mid a_{0:i-1},\, o,\, p,\, \theta) \right]
\end{align}
Due to the substantial parameter size of contemporary VLA models and their AR prediction strategy, the action speed of robot is inherently limited.
\begin{figure*}[t]
\centering
\includegraphics[width=1.8\columnwidth]{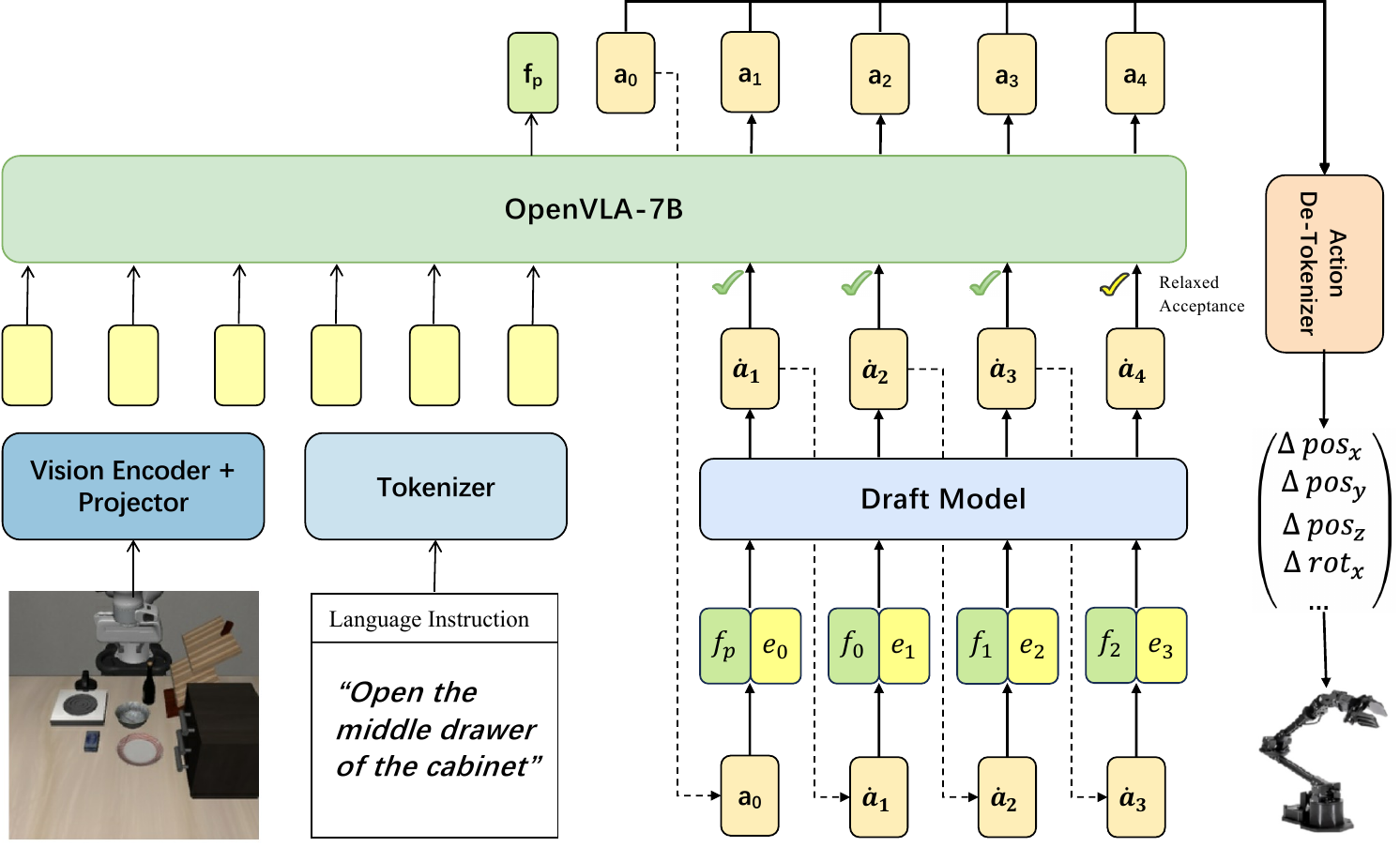}
\caption{The overall Spec-VLA framework. The draft model predicts action tokens through AR decoding with the fused textual and visual features. During verification, a relaxed acceptance mechanism is adopted to broadly retain high-quality outputs. This mechanism allows synonym to be accepted, while maintaining the success rate of action generation, achieving optimal balance between caption accuracy and efficiency.}
\end{figure*}
\begin{figure}[h]
\centering
\includegraphics[width=1\columnwidth]{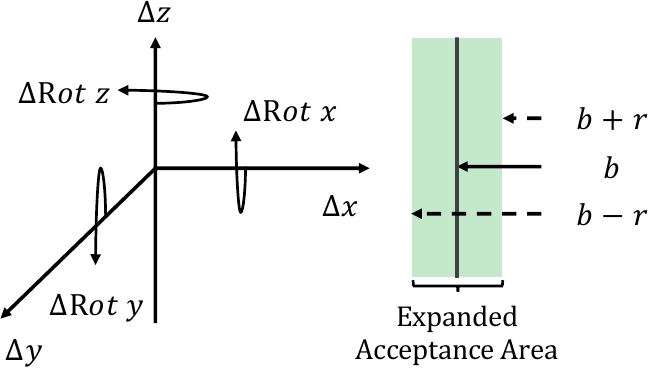}
\caption{Illustration of relaxation of the acceptance criteria. Instead of strictly accepting the predicted verify token $a_i$, the verification model $M_V$ accept action tokens within a predefined margin.}
\end{figure}
\subsection{Speculative Decoding Framework}
The SD framework utilizes an efficient \textbf{draft model}~$M_D$ to produce initial draft tokens and concurrently verifies these tokens using a  \textbf{verification model}~$M_V$. The Eagle framework~\cite{li2024eagle} incorporates a Llama layer as the draft model, which predicts multiple draft tokens $a_i$ autoregressively, conditioned on the previous draft token states $\hat{a}_{t+1:i-1}$, hidden states and token embeddings from verification model $f_{1:t}$ and  $e_{0:t}$. It is noteworthy that the output states of the draft tokens also assist in calculations, and for the sake of simplicity, we use the notation $\hat{a}_{t+1:i-1}$ to denote both embeddings and hidden features. 
\begin{align}
\label{Formulation_Drafting}
    \hat{a}_{i} = M_{D}(f_{1:t},e_{0 :t},\hat{a}_{t+1:i-1})
\end{align}
During the verification phase, the verification model $M_V$ ensures the generation quality of the draft tokens by correcting the mispredicted tokens from the draft model. When conducting a greedy search, the draft token $\dot{a}_i$ will be accepted only if it strictly matches the token $a_i$ predicted by the verification model.
\begin{align}
\begin{aligned}
 a_i &= M_V\bigl(a_1\sim \hat{a}_{i-1}
,\,p,\,\theta\bigr),\nonumber \\[-0.3em]
\end{aligned} \\
  \left\{
  \begin{aligned}
    &\text{Accept},            &&a_i==\hat a_i,\\
    &\text{Resample }\hat a_i=a_i,  &&a_i\neq\hat a_i.
  \end{aligned}
  \right.
\end{align}
Noticeably, the tokens subsequent to the first rejected token $a_{(i+1:L)}$ will be abandoned. Thus, the acceptance length is critical for the SD system as it determines the number of tokens to be predicted in a single forward pass.
\section{Spec-VLA Framework}
\begin{table*}[t]
\begin{center}
\begin{tabular}{l|c|ccc|ccc}
\toprule
\multirow{2}*{Dataset} & \multicolumn{1}{|c|}{AR} & \multicolumn{3}{|c|}{Spec-VLA} & \multicolumn{3}{|c}{Spec-VLA~(relaxed)} \\
\cmidrule{2-8}
               & SR & Length & Speedup & SR & Length  & Speedup  & SR\\
\midrule
LIBERO-Goal & \textbf{78.0}\% & 2.04 & $1.09\times$ & 74.2\% & \textbf{2.94} & $\mathbf{1.42\times}$ & 74.4\% \\
LIBERO-Object & \textbf{89.0\%} & 1.75 & $1.15\times$ & \textbf{89.0\%} & \textbf{2.38} & $\mathbf{1.38\times}$ & 85.0\% \\
LIBERO-Spatial & 85.0\% & 1.59 & $1.08\times$ & 83.8\% & \textbf{2.14} & $\mathbf{1.28\times}$ & \textbf{85.8\%} \\
LIBERO-Long & 52.0\% & 1.67 & $1.13\times$ & 50.8\% & \textbf{2.10} & $\mathbf{1.22\times}$ & \textbf{55.0\%}\\
\bottomrule
\end{tabular}
\end{center}
\caption{Experimental results of the Spec-VLA framework on the LIBERO-Goal, Object, Spatial, Long dataset. `SR' denotes the Success Rate of the control policy, `Length' indicates the number of tokens predicted in each forward pass, and `Speedup' reflects the generation speed as compared to the AR baseline.}
\label{main_result}
\end{table*}
In this section, we provide a detailed description of the Spec-VLA framework and our exploration of the adaptation of speculative execution for VLA prediction tasks.
\subsection{Overall Framework}
The Spec-VLA framework incorporates a Llama decoder layer~\cite{touvron2023llama} as its draft generator model. It incorporates a linear layer to integrate feature-level and token-level loss data effectively. During the prefill stage, the draft generator receives hidden states from the verification model, alongside textual and visual embeddings from the textual tokenizer and visual encoder, respectively. Mirroring the OpenVLA model, the visual embeddings $e_v$ and textual embeddings $e_T$ are concatenated,  collectively providing the feature-level information for the draft model.
\begin{align}
    \hat{a}_i = M_D(f_{1:t},\text{concat}(e_v,e_p),\hat{a}_{t+1:i-1}))
\end{align}

In the draft prediction phase, the draft generator model predicts the action token $a_i$ conditioned on previous hidden states, embeddings, and action tokens. We employ the dynamic draft tree strategy of Eagle-2~\cite{li2024eagle}, where the Top-K predictions from the draft generator $M_D$ are recorded and subsequently form a tree structure with multiple paths. These paths are then verified in parallel by the verification model.
\subsection{Problem by Direct Application}
However, directly implementing the SD framework yields only minor speed improvements, from $1.08\times$ to $1.15\times$~(as shown in Table \ref{main_result}). Surprisingly, in the VLA prediction task, the draft generator models fail to predict the initial draft tokens in about half of the samples~(refer to Table \ref{tab:proportions}). 

In natural language generation tasks, the draft generator of the SD framework typically produces common words and punctuation. Conversely, the VLA draft model must understand multiple modalities and predict robotic motions in VLA prediction tasks. Intuitively, the VLA prediction task poses a greater complexity for the draft generator than language generation. 

Moreover, VLA models such as OpenVLA and RT-2 incorporate greedy decoding during the drafting phase. This setup requires an exact match between draft tokens $\dot{a}_i$ and the verification model's predictions $a_i$. Often, allowing for synonym tokens could improve generation speed without compromising quality. Building upon prior research, we propose relaxing the acceptance criteria within the Spec-VLA framework by allowing the acceptance of top-k similar tokens in the action space.
\begin{figure*}[t]
\centering
\includegraphics[width=2\columnwidth]{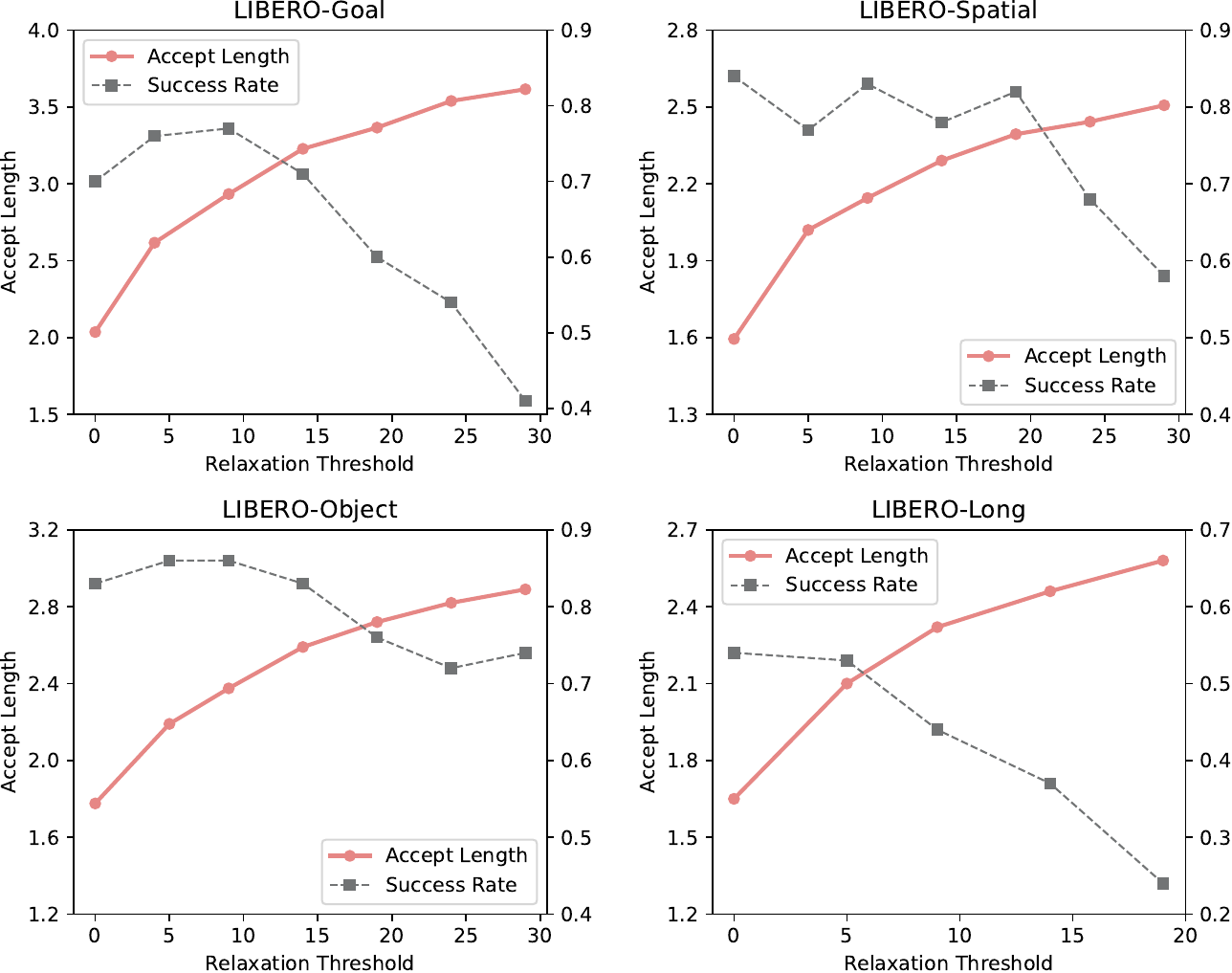}
\caption{Acceptance Length and Success Rate of the Spec-VLA framework on the LIBERO-Goal, LIBERO-Spatial, LIBERO-Object, and LIBERO-Long datasets. An increase in the Relaxation Threshold shows a minor impact on the Success Rate while significantly boosting the Acceptance Length. }
\label{Analysis_threshold}
\end{figure*}
\subsection{Relaxation of Acceptance}

We introduce the Relaxation Threshold $r$ to facilitate acceptance relaxation, quantifying the permissible distance between the draft action token $\hat{a}_i$ and the predicted action token $a_i$. The draft token $\hat{a}_i$ will be accepted if the distance $D$ between $\hat{a}_i$ and $a_i$ is not larger than threshold $r$. 
\begin{align}
\begin{aligned}
 a_i &= M_V\bigl(a_1\sim \hat{a}_{i-1}
,\,p,\,\theta\bigr),\nonumber \\[-0.3em]
\end{aligned} \\
  \left\{
  \begin{aligned}
    &\text{Accept},            &&D(a_i, \hat{a}_i)\leq r\\
    &\text{Resample }\hat a_i=a_i,  &&D(a_i, \hat{a}_i) > r .
  \end{aligned}
  \right.
\end{align}

VLA models, notably OpenVLA and RT-2, discretize continuous dimensions into 256 bins and map them to 256 action tokens to predict action sequences. The VLA token representation inherently provides information on token similarity, where the distance between tokens can be directly inferred from the absolute difference between bin IDs. For instance, the token $a$ represents bin $b$ and the token $\hat{a}$ represents bin $\hat{b}$, the distance between $a$ and $\hat{a}$ can be directly determined by the absolute difference between bin IDs $b$ and $\hat{b}$. The token acceptance area will be widened from strictly $b$ to $\hat{b} \in (b-r,b+r)$, enabling the acceptance of the top $2\times r$ similar tokens.

By utilizing this characteristic, our proposed method eliminates the need for additional token similarity calculations from token embeddings, introducing virtually no computational overhead.


\section{Experiment}
\subsection{Main Result}
Following OpenVLA, we evaluated the Spec-VLA framework on the LIBERO simulation benchmark~\cite{liu2023libero}.  We utilized four task suites: LIBERO-Object, LIBERO-Spatial, LIBERO-Goal, and LIBERO-Long, each providing 10 tasks and 500 expert demonstrations. We employed the finetuned OpenVLA as the verification model and used this model to regenerate the dataset for training the draft model. We conducted 50 trials on each task with our SD frameworks for testing scenarios. The training was completed in 6 hours using $4\times$ Tesla A100~(80G) GPU, with a batch size of 16. We inherent the implementation of draft model structure and tree decoding mechenism from Eagle-2~\cite{li2024eagle}. For tree-decoding, we set the maximum nodes to 50, tree depth to 4, and used the top 8 tokens to construct the draft tree. Drawing on prior works in SD~\cite{jang2024lantern}, we report the number of tokens predicted in each forward pass and speedup compared to AR decoding.

The main results are reported in Table \ref{main_result}. Firstly, the results validate the effectiveness of the SD framework in VLA prediction scenarios. Applying the Eagle framework achieves an acceleration ratio ranging from $1.08\times$ to $1.15\times$ without sacrificing generation quality. Secondly, the relaxed acceptance mechanism further enhances the generation speed of the SD framework, increasing the acceptance length by $25\%$ to $44\%$, demonstrating the potential for developing specialized SD mechanisms in the VLA scenario.
\begin{table*}[t]
\centering
\begin{tabular}{l|c|c|c|c|c|c|c}
\toprule
\multirow{2}{*}{Dataset} & \multirow{2}{*}{Relaxed} & \multicolumn{6}{c}{Acceptance Length} \\
\cmidrule(l){3-8}
 & & 0 & 1 & 2 & 3 & 4 & 5 \\
\midrule
\multirow{2}{*}{Libero-Goal} 
    & $\times$    & 50.24\% & 33.28\% & 13.96\% & 2.23\% & 0.19\% & 0.00\% \\
    & $\checkmark$ & 23.01\% & 18.98\% & 39.99\% & 15.41\% & 2.53\% & 0.08\% \\
\midrule  
\multirow{2}{*}{Libero-Object} 
    & $\times$    & 47.93\% & 34.72\% & 12.72\% & 4.07\% & 0.56\% & 0.00\% \\
    & $\checkmark$ & 28.23\% & 37.62\% & 17.29\% & 10.11\% & 6.22\% & 0.53\% \\
\midrule  
\multirow{2}{*}{Libero-Spatial} 
    & $\times$    & 55.96\% & 31.52\% & 9.90\% & 2.46\% & 0.16\% & 0.00\% \\
    & $\checkmark$ & 37.07\% & 33.52\% & 19.15\% & 8.23\% & 1.99\% & 0.03\% \\
\midrule  
\multirow{2}{*}{Libero-Long} 
    & $\times$    & 55.08\% & 28.77\% & 11.30\% & 4.30\% & 0.50\% & 0.05\% \\
    & $\checkmark$ & 42.39\% & 28.57\% & 16.65\% & 8.84\% & 3.35\% & 0.20\% \\
\bottomrule
\end{tabular}
\caption{Acceptance length distribution on the LIBERO-Goal, LIBERO-Object, LIBERO-Spatial, and LIBERO-Long datasets under non-relaxed and relaxed settings. Each row reports the proportion of trials that succeeded with a specific acceptance length. The threshold for relaxation is 9 for LIBERO-Goal, LIBERO-Object, and LIBERO-Spatial, and 5 for LIBERO-Long. }

\label{tab:proportions}
\end{table*}
\subsection{Ablations on Relaxation Threshold}
%
This section further analyzes the relationship between the relaxation threshold, success rate, and acceptance length (as shown in Figure \ref{Analysis_threshold}). We conducted analyses on the LIBERO-Goal, LIBERO-Spatial, LIBERO-Object, and LIBERO-Long benchmarks, each containing 10 tasks, with 10 trials performed for each task. We tested starting from a relaxation threshold of 0, which corresponds to strict matching acceptance.

First, Relaxation of acceptance criteria effectively enhances the acceptance length, boosting the generation speed of the VLA models. The increase in relaxation distance can enhance the acceptance length by $50\%$ to $70\%$ across various datasets. Moreover, we surprisingly found that the OpenVLA model displays high robustness on the LIBERO-Goal, LIBERO-Object, and LIBERO-Spatial datasets. The relaxation threshold could be relaxed from 5 to 9 without sacrificing the success rate of the VLA model.

Additionally, the better a model performs in a scenario, the larger the relaxation threshold it can tolerate. In the LIBERO-Long dataset, the success rate drops significantly when the relaxation threshold exceeds 5. However, in LIBERO-Goal, the success rate remains stable even with the relaxation threshold set to 15. This analysis verifies the effectiveness of our proposed relaxed acceptance strategy and also highlights the high potential for speculative execution within the VLA framework.
\begin{table*}[t]
\centering
\begin{tabular}{l|c|c|c|c|c|c|c}
\toprule
\multirow{2}{*}{Dataset} & \multirow{2}{*}{Relaxed} & \multicolumn{6}{c}{Position} \\
\cmidrule(l){3-8}
 & & 0 & 1 & 2 & 3 & 4 & 5 \\
\midrule
\multirow{2}{*}{Libero-Goal} 
    & $\times$ & 0.47 & 0.30 & 0.73 & 0.78 & 1.13 & 0.98 \\
    & $\checkmark$ & 1.44 & 1.36 & 2.18 & 2.13 & 1.76 & 0.98 \\
\midrule 
\multirow{2}{*}{Libero-Object} 
    & $\times$ & 0.60 & 0.64 & 1.02 & 0.67 & 0.88 & 0.99 \\
    & $\checkmark$ & 1.39 & 2.09 & 1.66 & 1.46 & 1.40 & 0.99 \\
\midrule 
\multirow{2}{*}{Libero-Spatial} 
    & $\times$ & 0.36 & 0.50 & 0.88 & 0.72 & 0.70 & 0.96 \\
    & $\checkmark$ & 0.89 & 1.39 & 1.55 & 1.56 & 1.20 & 0.98 \\
\midrule 
\multirow{2}{*}{Libero-Long} 
    & $\times$ & 0.79 & 0.42 & 1.19 & 0.87 & 0.65 & 0.64 \\
    & $\checkmark$ & 1.27 & 0.94 & 1.70 & 1.35 & 1.12 & 0.72 \\
\bottomrule
\end{tabular}
\caption{Average acceptance lengths at each position (0–5) on the LIBERO-Goal, LIBERO-Object, LIBERO-Spatial, and LIBERO-Long datasets under non-relaxed and relaxed conditions. Each entry reports the average acceptance length observed at the given token position. The relaxed setting is consistent with Table~\ref{tab:proportions}.}
\label{tab:positions}
\end{table*}
\begin{figure*}[t]
\centering
\includegraphics[width=2\columnwidth]{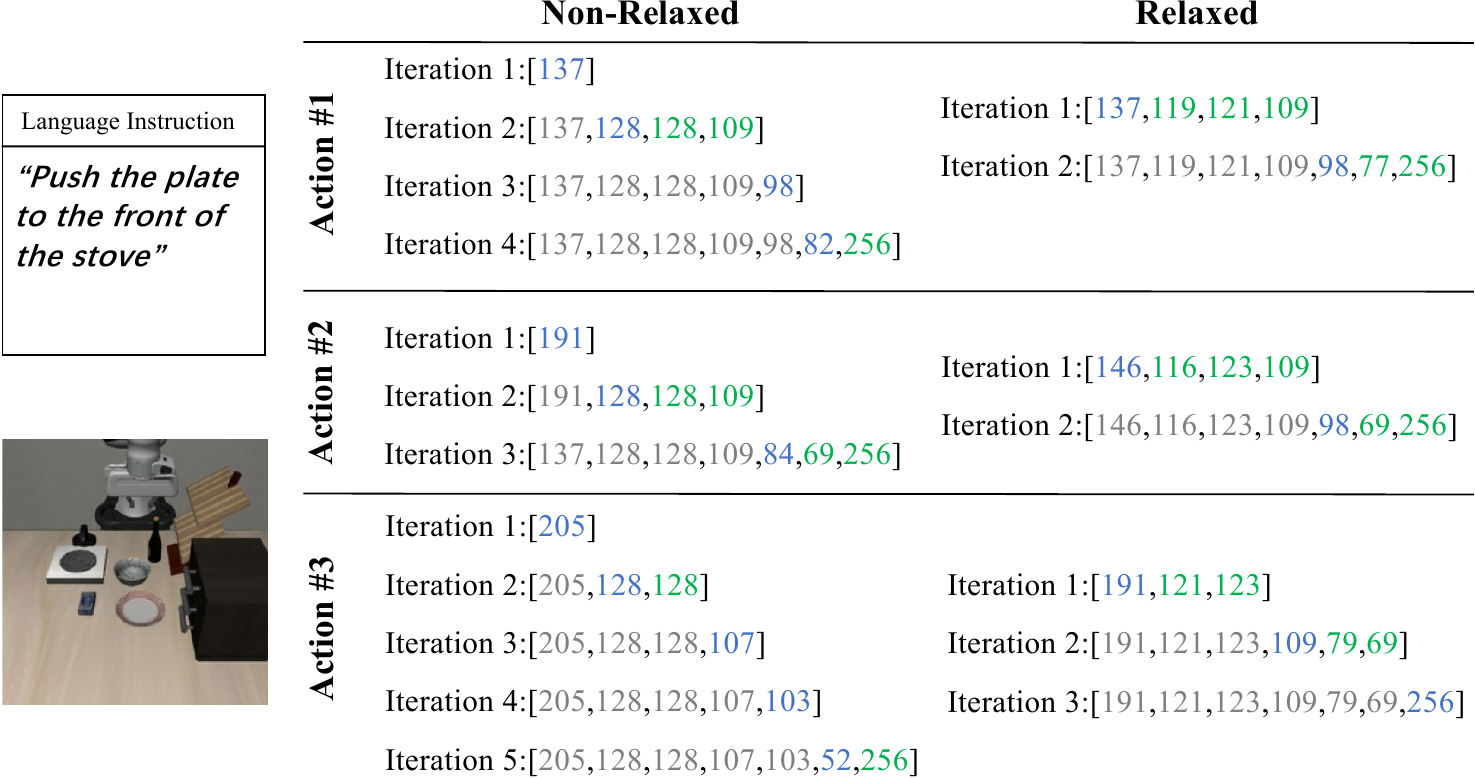}
\caption{Illustration of action sequence generation cases under non-relaxed and relaxed acceptance conditions in the Spec-VLA framework. Three representative action trajectories are juxtaposed for systematic comparison across both conditions. Gray denotes context tokens. Blue represents verification model outputs. Green indicates draft model outputs.}
\label{fig:case_study}
\end{figure*}
 \section{Analysis}
 This section provides an analysis of the Spec-VLA framework under non-relaxed and relaxed acceptance conditions, focusing on acceptance length distribution patterns and prediction performance across distinct action tokens on four benchmark datasets~(Libero-Goal, Libero-Object, Libero-Spatial, and Libero-Long). Consider that the verification model invariably emits an accept length of 1, which carries no discriminative information; our analysis here considers only the accept lengths produced by the draft model. Once verification outputs are excluded, the minimum accept length becomes 0 (indicating no speculative tokens were accepted), so the average accept length on each position can legitimately fall below 1.
 \subsection{Acceptance Length Proportion}
 \label{sec:Table 2}
 Table~\ref{tab:proportions} quantifies the distribution of acceptance lengths (0–5) under the Spec-VLA framework, comparing non-relaxed and relaxed conditions across four datasets. The data reveals a distinct trend: non-relaxed acceptance disproportionately favors shorter sequences~(lengths 0–1), with proportions sharply declining for longer lengths~(2–5), whereas relaxed acceptance exhibits a more balanced distribution. The dominance of short sequences under non-relaxed conditions~(e.g., 50.24\% at length 0) highlights a critical inefficiency: models prioritize `safe' short predictions to avoid constraint violations. This artificially low conversion rate for longer sequences implies that strict constraints act as a bottleneck preventing the model from predicting longer action sequence. The most pronounced contrast occurs in Libero-Object at length 4: non-relaxed acceptance plummets to 0.56\% versus 6.22\% under relaxed conditions—an 11-fold relative increase. Similarly, Libero-Long exhibits dramatic divergence at length 4 (0.50\% vs. 3.35\%, 6.7× improvement) and Libero-Spatial at length 3 (2.46\% vs. 8.23\%, 3.3× improvement). Even at maximum length 5, relaxed acceptance achieves non-zero proportions (e.g., 0.53\% for relaxed in Libero-Object vs. 0\% non-relaxed).
These disparities highlight a critical limitation of strict constraints: they disproportionately penalize longer sequences. Relaxation alleviates this by allowing semantically compatible draft tokens to be accepted, thereby increasing sequence diversity without compromising task success rates. 
\subsection{Acceptance Length on Multiple Positions}
We perform further analysis to evaluate the acceptance length in each starting position. 
As shown in Table~\ref{tab:positions}, relaxed acceptance consistently achieves longer average lengths than non-relaxed acceptance across all positions. For Libero-Object, acceptance length at position 1 surges from 0.64~(non-relaxed) to 2.09 under relaxation (3.3× improvement), reflecting reduced bias toward short-term predictions. Similarly, Libero-Goal shows a 3.1× increase at position 0 (0.47 → 1.44), highlighting the model’s willingness to explore initial reasoning steps when constraints are loosened. Libero-Spatial also exhibits a 2.2× gain at position 3 (0.72 → 1.56), revealing that relaxation mitigates premature truncation of valid action sequences, whereas relaxation balances risk and exploration to unlock the potential for longer action sequence generation. These results align with findings in Table~\ref{tab:proportions}.
\subsection{Case Study}
This section provides a representative case to show the effectiveness of our proposed relaxation of acceptance method. As shown in Figure~\ref{fig:case_study}, under the strict verification model~(Non-Relaxed), the series appends only those candidate tokens that satisfy a stringent acceptability threshold, resulting in a gradual accretion of the action sequence. For instance, Action 1 extends from a solitary context token [137] to the fully verified sequence [137, 128, 128, 109, 98, 82, 256] over four iterative refinement steps. In contrast, by relaxed acceptance, the relaxed criterion admits a broader spectrum of draft proposals at an earlier stage; Action 1 already incorporates the tokens [119, 121, 109] in its initial iteration and further augments this set with [98, 77, 256] in the second iteration. The same pattern holds for the other cases. Action 3, for example, reaches the whole sequence [191, 121, 123, 109, 79, 69, 256] in only three iterations under the relaxed acceptance, whereas the non-relaxed acceptance requires five iterations. These results show that relaxing the acceptance threshold significantly reduces the number of iterations needed for plan generation while still preserving the quality of the final action sequences. This relaxation also accelerates the action sequence completion process, reducing the number of iterations without compromising functional validity.
\section{Conclusion}
In this study, we explore the application of the SD framework in VLA prediction tasks. We propose Spec-VLA, which enhances the Eagle framework for VLA predictions. To further boost the generation speed of the framework, we introduce the distance-sensitive relaxation of the acceptance strategy, which utilizes the token representation of VLA models to effectively identify the distance between action tokens and relax the acceptance threshold within the SD framework. Experimental results verify the effectiveness of the Spec-VLA framework, where the relaxation of acceptance criteria further boosts the acceptance length by 25\% to 44\% without compromising the success rate. Our findings on the relaxation of acceptance show high robustness of the VLA models, demonstrating the potential of speculative systems in the VLA prediction domain.
\section*{Limitations}
This work explores speculative decoding in VLA prediction tasks. Due to time and resource constraints, experiments were not conducted in real-world robotic settings. Additionally, due to limitations of the verification model, Action Chunking was not explored. Future work could incorporate additional methodologies into the SD framework for VLA models.
\section*{Acknowledgements}
This work was supported in part by the Science and Technology Development Fund of Macau SAR (Grant No. FDCT/0007/2024/AKP), the Science and Technology Development Fund of Macau SAR (Grant No. FDCT/0070/2022/AMJ, China Strategic Scientific and Technological Innovation Cooperation Project Grant No. 2022YFE0204900), the Science and Technology Development Fund of Macau SAR (Grant No. FDCT/060/2022/AFJ, National Natural Science Foundation of China Grant No. 62261160648), the UM and UMDF (Grant Nos. MYRG-GRG2023-00006-FST-UMDF, MYRG-GRG2024-00165-FST-UMDF, EF2024-00185-FST), and the National Natural Science Foundation of China (Grant No. 62266013).

\bibliography{custom}
\clearpage
\appendix
\label{sec:appendix}
\section{Parameter Settings}
\begin{table}[t]
\centering
\footnotesize
\setlength{\tabcolsep}{2pt}
\begin{tabular}{lc}
\toprule
{Parameter} & {Value} \\
\midrule
Learning Rate & 5e-5 \\
Batch Size & 16 \\
Warmup Steps & 2000 \\
$p_w$ & 0.1 \\
$v_w$ & 1.0 \\
Gradiant Clipping & 0.5 \\
\midrule
Top-k & 8 \\
Tree Depth & 5\\
Max Nodes & 50\\
\bottomrule
\end{tabular}
\caption{Parameter Settings of Spec-VLA Framework.}
\label{specvla_parameter}
\label{tab:parameter_setting}
\end{table}
 This section details the parameter settings of the Spec-VLA model. Table \ref{specvla_parameter} presents the training and inference parameters of the Spec-VLA framework. For LIBERO-Goal, LIBERO-Spatial, and LIBERO-Object, the relaxation threshold is set at 9, while for LIBERO-Long, it is set at 5. The parameters $p_w$ and $v_w$ represent the weights of the Cross-Entropy loss and Regression loss, respectively, as implemented in the Eagle configuration~\cite{li2024eagle_o}. 
 \section{Spec-VLA Decoding Algorithm}
 To enhance the understanding of the decoding process within the Spec-VLA framework, we provide pseudocode that illustrates Spec-VLA decoding with relaxed acceptance, as outlined in Algorithm~\ref{alg:SD_relaxed_acception}. 

 \begin{algorithm*}[t]
\linespread{1.2}\selectfont
\caption{Spec-VLA Decoding}
\label{alg:SD_relaxed_acception}
\begin{algorithmic}[1]
\State Input: Prompt $p$, Observation $o$, Verification Model $M_V$, Draft Model $M_D$, Verification model hidden states $f_{1:t}$, Visual and textual embeddings $e_{0:t}$, Search Depth $d$, Target Length $L$, Relaxation Threshold $r$ 
  \State init $n \xleftarrow{} t$
  \While{$n < L$}
    \For{$i$ in \{1,...,d\}}
        \State Sample draft in AR manner $\hat{a}_{i}=M_D(f_{1:t},e_{0:t},\hat{a}_{t+1:i-1})$
    \EndFor
    \State Compute the reference token set $a_{t+1:t+1+d}$ in parallel:
    $a_i = M_V(\hat{a}_{t+1:i-1},o,p)$
    \For{$i$ in \{t+1,...,t+1+d\}}
         \If{D($a_i$,$\hat{a}_i$)$<=$r}
         \State Set $a_i \leftarrow  \hat{a}_i $
         \Else
         \State $a_i \leftarrow a_i$
         \State break
        \EndIf
    \EndFor
    \State if all drafts accepted, sample an extra token $a_{t+d+2}=M_V(a_{t+1:t+d+1},o,p)$
  \EndWhile
  \State \Return $a_{t+1:t+d+1}$ or $a_{t+1:t+d+2}$
\end{algorithmic}
\end{algorithm*}
\section{Accelleration for Quantilized Models}
To illustrate SD’s complementary potential, we explored further by combining Speculative Decoding with quantization.

We compared inference speeds for the OpenVLA model using int8, int4 quantization, and BF16 representations on the Tesla A100 (80G) GPU. We observed that the int8 and int4 quantization lead to decreased inference speed~(Table~\ref{quantilized_openvla}). This result may be attributed to the additional overhead incurred by the quantization operation, consistent with the analysis of OpenVLA~\cite{kim2024openvla}. Additionally, we accelerated the quantized model using the Spec-VLA framework, discovering that SD could speed up the quantized verification model. It achieves a significant speedup compared with AR decoding~(Table~\ref{quantilized_specvla}).

\begin{table}[t]
\begin{center}
\begin{tabular}{l|cc}
\toprule
 {Dataset}  & \multicolumn{2}{|c}{OpenVLA} \\
\cmidrule{2-3}
                & Precision & Speedup    \\
\midrule
LIBERO-Goal & bf16 & 1.00$\times$  \\
 & int8 & 0.24$\times$ \\
 & int4 & 0.61$\times$ \\
\midrule
LIBERO-Object & bf16 & 1$\times$  \\
 & int8 & 0.21$\times$ \\
 & int4 & 0.59$\times$ \\
\midrule
LIBERO-Spatial & bf16 & 1$\times$ \\
 & int8 & 0.23$\times$ \\
 & int4 & 0.55$\times$ \\
\midrule
LIBERO-Long & bf16 & 1$\times$ \\
 & int8 & 0.23$\times$ \\
 & int4 & 0.57$\times$ \\
\bottomrule
\end{tabular}
\end{center}
\caption{Speedup of the quantilized OpenVLA model on the LIBERO-Goal, Object, Spatial, Long dataset. The `Precision' shows the quantization precision, and `Speedup' reflects the generation speed compared to the bf16 baseline.}
\label{quantilized_openvla}
\end{table}
\begin{table}[t]
\begin{center}
\begin{tabular}{l|cc}
\toprule
 {Dataset}  & \multicolumn{2}{|c}{SpecVLA} \\
\cmidrule{2-3}
                & Precision & Speedup    \\
\midrule
LIBERO-Goal & bf16 & 1.42$\times$  \\
 & int8 & 1.61$\times$ \\
 & int4 & 1.34$\times$ \\
\midrule
LIBERO-Object & bf16 & 1.38$\times$  \\
 & int8 & 1.41$\times$ \\
 & int4 & 1.33$\times$ \\
\midrule
LIBERO-Spatial & bf16 & 1.28$\times$ \\
 & int8 & 1.31$\times$ \\
 & int4 & 1.29$\times$ \\
\midrule
LIBERO-Long & bf16 & 1.22$\times$ \\
 & int8 & 1.32$\times$ \\
 & int4 & 1.15$\times$ \\
\bottomrule
\end{tabular}
\end{center}
\caption{Speedup of the quantilized SpecVLA framework on the LIBERO-Goal, Object, Spatial, Long dataset. The `Precision' shows the quantilization precision, `Speedup' reflects the generation speed as compared to the OpenVLA AR baseline.}
\label{quantilized_specvla}
\end{table}
\end{document}